\documentclass{article}
\usepackage[round,authoryear,comma]{natbib}
\usepackage[final]{neurips_2019}
\usepackage[utf8]{inputenc} 
\usepackage[T1]{fontenc}    
\usepackage{hyperref}       
\usepackage{url}            
\usepackage{booktabs}       
\usepackage{amsfonts}       
\usepackage{nicefrac}       
\usepackage{microtype}      
\usepackage{color,soul}
\usepackage{amsmath , amssymb , amsthm,fancyhdr,graphicx} 
\usepackage[utf8]{ inputenc}
\usepackage[ruled,vlined]{algorithm2e}
\usepackage{multirow}
\usepackage{colortbl}
\definecolor{kugray5}{RGB}{224,224,224}
\title{Potential adversarial samples for white-box attacks}

\author{%
  Amir Nazemi \\
  Department of Systems Design Engineering\\
  University of Waterloo\\
  Waterloo, Ontario, Canada.\\
  \texttt{anazemi@uwaterloo.ca}
  \And
  Paul Fieguth \\
  Department of Systems Design Engineering \\
  University of Waterloo\\
  Waterloo, Ontario, Canada.\\
  \texttt{paul.fieguth@uwaterloo.ca} 
}

\begin{document}

\maketitle

\begin{abstract}
Deep convolutional neural networks can be highly vulnerable to small perturbations of their inputs, potentially a major issue or limitation on system robustness when using deep networks as classifiers. In this paper we propose a low-cost method to explore marginal sample data near trained classifier decision boundaries, thus identifying  potential adversarial samples.  By finding such adversarial samples it is possible to reduce the search space of adversarial attack algorithms while keeping a reasonable successful perturbation rate. In our developed strategy, the potential adversarial samples represent only $61\%$ of the test data, but in fact cover more than $82\%$ of the adversarial samples produced by iFGSM and $92\%$ of the adversarial samples successfully perturbed by DeepFool on CIFAR10. 
\end{abstract}
\vspace{-0.2cm}
\section{Introduction}
\vspace{-0.3cm}

Deep Convolutional Neural Networks (DCNNs) have seized the attention of researchers throughout computer science and computer vision, with its usage nearly ubiquitous. Very troubling, however, is that DCNN models have been shown \cite{szegedy2013intriguing} to be sensitive to small perturbations in the input data, even perturbations so subtle that the human eye cannot distinguish the perturbed image from the original, and this, disturbingly, with DCNNs which otherwise were believed / tested to have high accuracy and robustness.

It is at or near to the classification boundary where a given DCNN typically has problems with the samples near the decision boundary, precisely because standard DCNN loss functions do not specifically consider any aspects of the decision boundaries, such as margin of separation between classes.
In this paper, we propose a low-cost method to determine the samples near the decision boundaries and by selecting those samples, the search space of adversarial methods will be smaller and assists in finding samples for potential adversarial attack, in order to more thoroughly be able to train for and validate algorithm robustness.
Our proposed method find samples near the decision boundaries using SVM and selecting its support vectors, on the basis of having full access to the target model, as in standard  \text{\em{white-box}} attacks.
\vspace{-0.2cm}

\section{Related work}
\vspace{-0.3cm}
Adversarial attacks are essentially a means of assessing network robustness.  Broadly, there are two different types of attack: \text{\em{white-box}} and \text{\em{black-box}} \cite{liu2016delving,chen2017zoo}. In \text{\em{white-box}} attacks an attacker has full access to the network model's parameters, whereas in \text{\em{black-box}} attacks the attacker has access only to network inputs and outputs, but not to any internal parameters. 
In this paper two \text{\em{white-box}} 
methods, iterative-Fast Gradient Sign Method (iFGSM) \cite{kurakin2016adversarial} and DeepFool \cite{moosavi2016deepfool}, are utilized.

Support Vector Machines (SVMs) as a conventional machine learning method have been evaluated against adversarial attacks \cite{biggio2011support,zhou2012adversarial,biggio2013evasion}; however, with the emergence of deep learning the focus of adversarial research has moved towards DCNNs.  It has shown that using SVM together with a DCNN achieves a higher accuracy than DCNNs alone on some datasets~\cite{tang2013deep}, suggesting that there are limitations or drawbacks with the classifier part of DCNNs. In \cite{kim2015deep}, the authors showed that the generalization problem of a DCNN can be handled by support vector data description (SVDD), a type of SVM.
Liang et al.~\cite{liang2017soft} proposed a soft margin softmax loss to improve the discriminative power of DCNN, however the DCNN problems go beyond softmax layer. Elsayed et al.~\cite{elsayed2018large} proposed a large margin loss function for DCNNs that creates a model which not only has a higher accuracy on the evaluated datasets but also is more robust against adversarial perturbations. Khoury et al.~\cite{khoury2018geometry} proposed a geometric framework to investigate geometric properties of adversarial examples. They argued that the adversarial samples are the consequence of learning the decision boundaries on a low-dimensional manifold, which is not sufficiently generalized to the actual manifold of data. Jiang et al.~\cite{jiang2018predicting} proposed a metric for calculating the generalization gap of a DCNN model based on the marginal distribution at some layers of the network.

\vspace{-0.2cm}
\subsection{iFGSM}
\vspace{-0.2cm}

The Fast Gradient Sign Method (FGSM) was first proposed by Goodfellow et al. \cite{goodfellow2014explaining} for generating $X_{adv}$ from an input image $X$. It uses the sign of the gradient ($\nabla$) of the loss function $l()$ of the network:
\begin{equation}
    X_{adv} = X + \epsilon~\text{\em sign}(\nabla_X l(X,Y_{\text{\em true}}))
\end{equation}
Here, $\epsilon$ is a positive constant and determines the size of perturbation. A more powerful variation of FGSM is iterative-FGSM (iFGSM) \cite{kurakin2016adversarial}, whereby  FGSM is undertaken in $n$ iterations with a smaller perturbation $\alpha = \frac{{\epsilon}}{n}$ in each step:
\begin{equation}
        X^{t+1} = X^t + \alpha~\text{\em sign}(\nabla_X l(X^t,Y_{\text{\em true}})) \qquad
        X_{adv} =X^n
\end{equation}
In iFGSM, the perturbation is bounded by ~$l_\infty \text{\em norm}$ to $\epsilon$, meaning that the maximum perturbation for each input pixel $x$ is itself bounded by $\epsilon$.
\vspace{-0.2cm}
\subsection{DeepFool}
\vspace{-0.2cm}
DeepFool~\cite{moosavi2016deepfool} is an accurate approach for constructing adversarial examples based on the distance of samples to the closest decision boundary. Deepfool tries to find the minimum perturbation $r$ that changes the output of function $f()$ if added to input $x$. The algorithm assumes that $f()$ is an affine binary classification function and estimates the perpendicular distance and direction from the input $x$ to the decision boundary $f(x)=0$.  The calculated perturbation $r$ is multiplied by a constant $1 + \zeta$ to ensure that $x$ crosses the decision boundary, resulting in a net  perturbation of $\delta$:
\begin{equation}
        \delta = r \cdot (1+\zeta)
        \qquad
        X_{adv} = X + \delta 
\end{equation}

 
    
   
\vspace{-0.2cm}
\section{Methodology}
\vspace{-0.3cm}

In machine learning a shallow model like SVM tries to find a decision boundary that has the largest margin of separation between classes. The resulting optimization problem is tractable when the complexity of the data is low; however, for problems of very high dimensionality (common in image-related classification) the complexity can be problematically high, so we desire an alternative strategy for finding marginal samples.
In contrast, deep models handle complex data very well, however they cannot (in general) handle data which are near decision boundaries, so they are sensitive to small perturbations of those samples.  Since DCNN models do not explicitly model or penalize classification decision boundaries, and as a result they can be vulnerable to small, even infinitesimal, changes on such input samples. 
Our method proposes the following steps for finding Potential Adversarial Samples (PAS):
\begin{enumerate}
\item Based on loss function $l(X,Y_{\text{\em true}})$, model $f()$ is trained.  The test data $X$ are fed to the model, and the outputs of the last convolutional layer of correctly classified samples $X_f$ are saved.
\item A one-versus-rest SVM model is trained on the saved features of the previous step. The SVM is not overfitted, in which case its decision boundaries are close to those of $f()$.
\item The resulting support vectors of the trained SVM, which will be near to the SVM classification boundaries,  represent our potential adversarial samples.
\end{enumerate}
To assess the above, test samples are perturbed using two different adversarial attack algorithms, such as iFGSM and DeepFool, and the degree of commonality or overlap between our selected PAS and the adversarial samples will be measured. 
\vspace{-0.2cm}
\section{Experimental Setup}
\begin{figure}
\begin{tabular}{cc}
    \includegraphics[width=0.45\textwidth]{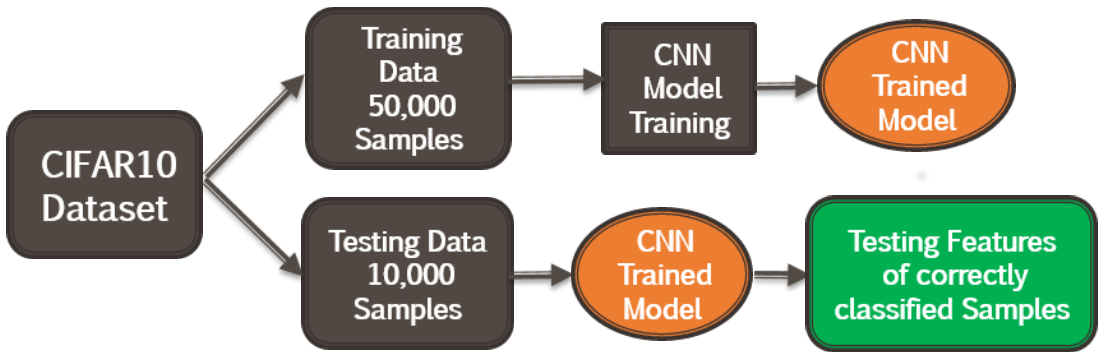} &
    \includegraphics[width=0.4\textwidth]{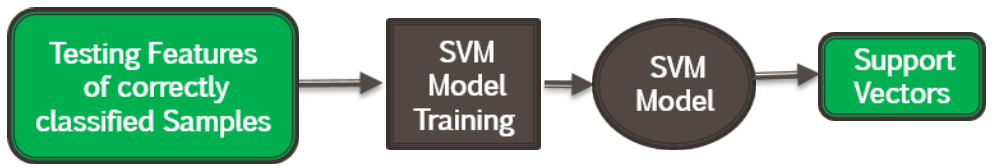}
    \\
    (A) & (B)
\end{tabular}
    \caption{The proposed framework to select support vectors as the PAS, here shown for CIFAR10. In (A) the test inputs transfer to the feature space; and in
    (B), the support vectors (PAS) are selected. Note that the framework selects support vectors from the test data.}
    \label{fig:methdology}
\end{figure}

\vspace{-0.3cm}
Two different convolutional neural network (CNN) models are used on two different datasets. LeNet~\cite{lecun2015lenet} is used as the simplest network, with two convolutional layers, applied to the MNIST dataset.  The VGG19 network, one of the biggest networks in the field of deep learning with 19 convolutional layers, was evaluated on CIFAR10. CIFAR10 has $60,000$ samples, of which $50,000$ samples are selected for training and the remainder for testing. MNIST has $60,000$ training samples and $10,000$ testing samples. Both datasets have 10 classes. As summarized in Figure~\ref{fig:methdology}, the PAS are selected among the testing data of the datasets which are correctly classified by the CNN models.

The SVM model requires the selection of parameter $C$, which tunes the SVM to be closer to soft- or hard-margin. In other words, smaller values of $C$ lead the SVM to allow some samples be in the marginal area; as $C$ is increased SVM becomes increasingly hard-margin, in which greater penalties are asserted for samples in the marginal area.
 
Other important experimental parameters are $\epsilon$ for iFGSM and $\delta$ for DeepFool, parameters which define the maximum amount of perturbation which the algorithm can employ on a sample in the $l_\infty-\text{\em norm}$ space.
\vspace{-0.2cm}
\section{Experimental Results}
\vspace{-0.3cm}

The LeNet network accuracy on the MNIST test set is $98.23\%$, meaning that $9823$ samples out of $10,000$ dastaset testing samples are correctly classified.  Attack algorithms seek to change the decision of the network on some of those $9823$ samples by perturbing the inputs. 

Table~\ref{table:Lenet-Mnist} shows the overlap between the PAS, which are selected by SVM models with different $C$ parameters, and the testing samples which were successfully perturbed by iFGSM with different $\epsilon$ and DeepFool with different $\delta$ values on the MNIST dataset, based on the LeNet network. As shown in the iFGSM part of  Table~\ref{table:Lenet-Mnist}, the PAS cover a minimum of $60\%$ of the adversarial samples while the search space of the model is reduced to less than $25\%$ of the input samples, with similar results in the bottom half of 
Table~\ref{table:Lenet-Mnist}
corresponding to DeepFool. The results suggest that the selection of PAS is only weakly dependent on the attack algorithm, and is instead related to the data distribution and the decision boundaries.

The MNIST dataset is too small and straightforward to meaningfully evaluate all the challenges of the proposed idea. So, a larger network model and a more complex dataset is needed.
The VGG19 network correctly classifies $9228$ samples of $10,000$ testing samples of CIFAR10. These $9228$ samples are the search space of attack algorithms.

The top (iFGSM) part of Table~\ref{table:VGG-CIFAR10-iFGSM} illustrates the number of samples which are common between the PAS and the adversarial samples of the iFGSM algorithm, as a function of $\epsilon$. The results show that we can reduce the search space of adversarial attack algorithms as we need and on that search space we will cover most of adversarial samples which are selected by the iFGSM algorithm. In comparison with MNIST dataset the variation of the CIFAR10 dataset is higher and in a result the number of adversarial is more than MNIST dataset. It means the decision boundary of the feature space is more complicated than the one of MNIST dataset. 
However, the CNN model here is VGG19 which is much bigger than LeNet and its features space is more complicated.

The DeepFool part of Table~\ref{table:VGG-CIFAR10-iFGSM} shows the same results for the DeepFool algorithm. As illustrated in that part of table, with the parameter value $\delta$ equal to or less than $0.0001$, the DeepFool algorithm can not perturb any test sample and in general, due to its step size, the algorithm finds less number of adversarial samples than those iFGSM finds in  a search area (neighborhood) in $l_\infty-norm$ space. However, the DeepFool can perturb any test sample in a fair number of iterations.

\begin{table}[t]
\scriptsize
\centering
\caption{The number of common samples between PAS and those which are successfully perturbed by iFGSM and DeepFool, here based on the LeNet network trained on the MNIST dataset. The coverage will clearly increase with the number of support vectors, which in turn depends on SVM parameter $C$.}
\begin{tabular}{|c|c|c|c|c|c|}
\hline
\arrayrulecolor{black}
\multicolumn{2}{|c|}{SVM $C$ Parameter}&5&2&1&0.5\\
\hline
\multicolumn{2}{|c|}{SVM Training Accuracy (\%)} &100&99.94&99.82&99.68\\
\hline
\multicolumn{2}{|c|}{Total Number of SV}                 &1239&1436&1763&2274\\    
\hline
\multicolumn{6}{|c|}{\bf iFGSM}\\
\hline
$\epsilon$&Num Adversarial samples found by iFGSM &\multicolumn{4}{c|}{Number of Samples in common}\\
\hline
0.0001&1&1&1&1&1\\
\hline
0.0005&2&2&2&2&2\\
\hline
0.001&5&4&5&5&5\\
\hline
0.005&26&25&26&26&26\\
\hline
0.01&69&67&68&69&69\\
\hline
0.05&500&429&457&471&493\\
\hline
0.1&1903&983&1108&1284&1482\\
\hline

\multicolumn{6}{|c|}{\bf DeepFool}\\
\hline
$\delta$ &Num Adversarial samples found by DeepFool &\multicolumn{4}{c|}{Number of Samples in common}\\
\hline
0.0001&1&1&1&1&1\\
\hline
0.0005&2&2&2&2&2\\
\hline
0.001&2&2&2&2&2\\
\hline
0.005&13&12&13&13&13\\
\hline
0.01&26&25&26&26&26\\
\hline
0.05&194&189&190&192&194\\
\hline
0.1&512&448&475&490&505\\
\hline
\end{tabular}
\label{table:Lenet-Mnist}
\end{table}


\begin{table}[t]
\scriptsize
\centering
\caption{As in Table~\ref{table:Lenet-Mnist}, but now for VGG19 trained on CIFAR10.  Boldfaced values are those which show a high coverage of adversarial examples but with relatively few support vectors.}
\begin{tabular}{|c|c|c|c|c|c|c|c|c|}
\hline
\arrayrulecolor{black}
\multicolumn{2}{|c|}{SVM $C$ Parameter} &500&100&10&1&0.5&0.1&0.035\\
\hline
\multicolumn{2}{|c|}{SVM Training Accuracy (\%)}&99.98&99.89&99.81&99.67&99.62&99.29&99\\
\hline
\multicolumn{2}{|c|}{Total Number of SV}&195 &290 &554&1273&1699&3508&5633\\                    
\hline
\multicolumn{9}{|c|}{\bf iFGSM}\\
\hline
$\epsilon$&Adversarial Samples found by iFGSM&\multicolumn{7}{c|}{Number of Samples in common}\\
\hline
0.0001&13&\text{\bf{11}}&13&13&13&13&13&13\\
\hline
0.0005&64&\text{\bf{60}}&64 &64&64&64&64&64\\
\hline
0.001&129&\text{{97}}&\bf{118}&127&129 &128&129&129\\
\hline
0.005&818&184&277&476&\text{\bf{747}}&748&818&818\\
\hline
0.01&1801&192&286&534&1126 &\text{{1288}}&1743&1799\\
\hline
0.05&4819&194&288&550&1267&1663&3255&\text{{4291}}\\
\hline
0.01&5636&194&288&551&1268&1666&3335&\text{{4603}}\\

\hline
\multicolumn{9}{|c|}{\bf DeepFool}\\
\hline
$\delta$ & \begin{tabular}{c} Adversarial samples found by DeepFool \end{tabular} &\multicolumn{7}{c|}{Number of Samples in common}\\
\hline
0.0001&0&\text{{0}}&0&0&0&0&0&0\\
\hline
0.0005&15&\text{\bf{13}}&15&15&15&15&15&15\\
\hline
0.001&26&\text{{24}}&\bf{26}&26&26&26&26&26\\
\hline
0.005&146&\text{{108}}&\bf{133}&145&146&146&146&146\\
\hline
0.01&314&150&\text{{212}}&\bf{292}&314&314&314&314\\
\hline
0.05&1615&193&287&539&1096&\text{\bf{1291}}&1572&1609\\
\hline
0.01&3667&194&288&550&1236&1603&\text{{2726}}&3376\\

\hline
\end{tabular}
\label{table:VGG-CIFAR10-iFGSM}
\end{table}

\vspace{-0.2cm}
\section{Discussion and Conclusion}
\vspace{-0.3cm}
This paper examined the test samples assigned on the basis of the support vectors of an SVM model, itself trained on the DCNN features, leading to a  proposed method which is a low-cost and fast method to identify potential adversarial samples.
It is  near-boundary samples upon which small perturbations are more likely to fool the network, and it is this understanding which helps adversarial attack algorithms to reduce their search space. 

To make an algorithm more defensive, a straightforward strategy is to add those adversarial examples into the training set and to retrain the model again.  This resistance to adversarial examples is encoded in the training phase, causing the model learn those examples and to be more
robust in dealing with those  examples. However, this defense mechanism is attack oriented, meaning that the model could be vulnerable to unseen adversarial attack methods whose samples are not considered in the training phase. In this case, using an ensemble classifier with a focus on potential adversarial samples can improve the robustness of the model. 

\small
\bibliographystyle{plainnat}
\bibliography{main}

\end{document}